# A Study of Deep CNN Model with Labeling Noise Based on Granular-ball Computing

Dawei Dai[1], Donggen Li[1], Zhiguo Zhuang[1]
1. Chongqing University of Posts and Telecommunications, Chongqing, China

## 1. Abstract

Label noise refers to the mislabeling of dataset samples. In the field of machine learning and deep learning, a small amount of label noise data may have a large negative impact on the final predicted classification results. Label noise can cause a series of problems such as decreased accuracy of model classification, inefficiency of classifier, and increased time and complexity, resulting in the inability to train an accurate and effective model. In supervised learning, the presence of noise can have a significant impact on decision making. It is found that excessive label noise in some commonly used algorithms, such as in the KNN nearest neighbor algorithm, can confuse the distribution of the data and thus affect the selection of the classifier K-values and the kernel classifier. Since many classifiers do not take label noise into account in the derivation of the loss function, including the loss functions of logistic regression, SVM, and AdaBoost, especially the AdaBoost iterative algorithm, whose core idea is to continuously increase the weight value of the misclassified samples, the weight of samples in many presence of label noise will be increased, leading to a decrease in model accuracy. In addition, the learning process of BP neural network and decision tree will also be affected by label noise. Therefore, solving the label noise problem is an important element of maintaining the robustness of the network model, which is of great practical significance.

Granular ball computing is an important modeling method developed in the field of granular computing in recent years, which is an efficient, robust and scalable learning method. In this paper, we pioneered a granular ball neural network algorithm model, which adopts the idea of multi-granular to filter label noise samples during model training, solving the current problem of model instability caused by label noise in the field of deep learning, greatly reducing the proportion of label noise in training samples and improving the robustness of neural network models.

## 2. Related work

In the field of machine learning and deep learning, a small amount of label noise data may have a large negative impact on the final predicted classification results[1]. Label noise can cause a series of problems such as decreased accuracy of model classification, inefficient classifiers and increased time and complexity, resulting in the inability to train accurate and effective models, and in supervised learning, the presence of noise can have a large impact on decision making[2]. Since many classifiers derive loss functions without taking label noise into account, including the loss functions of logistic regression, SVM, and AdaBoost, especially the AdaBoost iterative algorithm[3], whose core idea is to continuously increase the weight value of the misclassified samples, the weights are increased in many samples where label noise is present[4], leading to a decrease in model accuracy. In addition, the learning process of BP neural network[5] and decision tree[6] will also be affected by label noise. Therefore, solving the label noise problem is an important element of maintaining the robustness of network models and is of great practical importance[7].

Currently, the common ways of noise handling are roughly divided into two types of noise accommodation and noise filtering[8], the former is to weaken the influence of label noise on the model by building a robust algorithmic model that is less sensitive to label noise, although robust learning algorithms are usually only applicable to simple cases, designing special algorithms for label noise is necessary and there are some advances, such as label noise filtering of Bayes model[9], robust SVMs and others[10], but these methods are designed for specific loss functions and there is no general framework that includes multiple loss functions. The latter is done by algorithmically removing or correcting noisy samples from the data before returning the model for training[11]. When the proportion of labeled noise is high, removing all the noisy samples will drastically cut down the content of the dataset, which may lead to imbalance in the categories and insufficient information analysis of the data, so the corresponding labeled noise filtering classifier should be selected according to the actual situation of the dataset.

## 3. Thoughts

In the field of deep learning images, the attributes of samples will be represented as spatial vectors with little difference in distance between samples with similar features, while samples with different labels with different features are far apart, so that mislabeled sample points have significant outliers.

To filter out the label noise that destroys the robustness of the model, we add the process of granular ball clustering in front of the model classification layer. That is, after the feature vector is passed out by the neural network model, it is clustered into each different ball cluster by the grain ball clustering algorithm. According to the feature that similar samples are close together, each ball contains a large number of correctly labeled samples and a small number of incorrectly labeled samples. The labels of the granular balls are set to the labels of most samples within the balls, so the correctness of the granular sphere labels can be guaranteed, and then the center point of the granular balls is used as a new sample point to achieve filtering of noisy samples.

## 4. Method

*A. Granular Ball Neural Network Model*
(1) The granular ball model obtains the feature vectors, labels, and the corresponding data format conversions after convolutional pooling of the neural network, respectively.

(2) The sample data are injected into the granular ball model and clustered by iterative granular ball splitting. Granular ball clustering is based on the feature distribution of different samples clustered in different balls, according to the majority of sample points in the granular ball to determine the label value of the granualr ball, in order to neutralize the labeled noise samples and determine whether the proportion of the majority of labeled sample points reaches the set threshold, if the threshold is reached, stop the split, if not, continue to split into smaller granular balls, until the purity of all granualr balls meet the standard or only a single sample point for those that do not meet the threshold.

(3) Due to the nature of grain ball clustering, most of the label noise sample points will eventually be clustered into single-sample balls, so by clustering small samples into multiple grains and finally discarding single-sample balls, the proportion of label noise in the samples will be greatly reduced.

(4) The final obtained grain sphere centroids are used instead of all sample points put into the classifier for training.

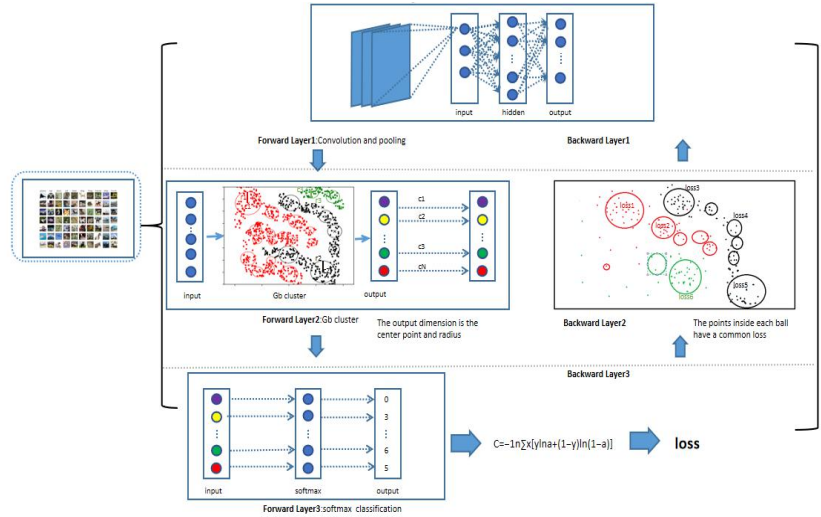

Figure 1. The overall network structure of our proposed model

## B. Technical points introduction

(1) Granualr ball centroid filtering label noise.

Reducing the proportion of label noise in the data set is the key to improving the robustness of the neural network model. Because the distribution of label noise sample points in space is outlier, the set of sample points inside the ball includes a certain number of noise points after clustering by the grain sphere, and these noise points will cause interference to the training of the model. Therefore, the granular ball neural network model calculates the center of each granule ball in the granular ball layer, determines the label of the granular ball according to the true labels of most samples in the granular ball, and then replaces all sample points in the granular ball with the characteristic value of the center point and the label of the granule ball.

(2) Multiple clustering filtering of single-sample balls

Through the experiment, we found that after each granular ball clustering, there will inevitably be a single sample point of the granular ball, and the proportion of label noise of these single point granular ball is significantly higher than the proportion of noise in the data set, so in order to get more and better center point samples, for these single sample point of the granular ball, we use multiple iterative clustering, and finally the remaining single sample point of the granular ball discarded.

(3) Back propagation of neural network gradients

Similar to the pooling layer of the neural network, because the granule ball neural network model in the granular ball layer filtered out part of the samples, so that the size of the feature map changes, if the mapsize of the granular ball before clustering is 512*32, after clustering becomes 128*32, only 128 gradients back to the granular ball layer, so this will make the gradient propagation interrupted. The back propagation of the granular ball layer is similar to the neural network pooling layer. Since the granular ball neural network model uses the centroid instead of the sample points in the ball, when the gradient is back propagated, the gradient passed to the granular ball layer is the gradient of the centroid, and the gradient should be mapped to all sample points in the previous layer to ensure the continuity of the neural network. Therefore, we first obtain the gradient vector of the centroid point passed back from the classifier layer, and then set the gradients of all sample points within the same granular ball to the gradient of the centroid point of that granular ball according to the information saved during forward propagation, encapsulate the gradient vector passed back, and continue to pass back to the upper layer.

(4) Experience replay mechanism

The experience replay method in the field of reinforcement learning breaks the correlation of sequences and can reuse the past experience to enable the model to be fitted effectively, because in the granular ball clustering, the large ball and the small ball have the same weight in the calculation of the loss, but the noise proportion of the large ball is much smaller than that of the small ball, so increasing the proportion of large balls in the training sample is a key step. Therefore, we store the data samples of each Batch and randomly select a certain number of larger balls to combine into a new Map to participate in training at the next Batch.

### C. *Granular ball neural network innovation and advantages.*
(1) Applying the idea of multiple granularity to the problem of dealing with deep learning label noise.
(2) After clustering and discarding a single sample point several times, the granular ball model replaces all sample points inside the ball with the granular ball centroids.
(3) Replacing all sample gradients inside the ball with the granular ball centroid gradients.
(4) The granular ball neural network model is independent of the neural network model and does not depend on any kind of loss function or network.

# 5. Experiments

The above section describes the effectiveness of the granular ball neural network model to solve the deep learning label noise problem from the theoretical principles. In order to combine theory with practice, we rewrite the underlying function when DataLoder is performed on the dataset, select a fixed proportion of noise samples for random label noise addition, and make the label noise sample points evenly distributed in the dataset. Meanwhile, this section uses the same dataset and validation method as in the paper to experimentally demonstrate the effectiveness of solving the problem of label noise for deep learning in this paper.

First, compaerd to the traditional convolutional neural network, we test the results of our model on the CIFAR-10 dataset, as shown in Table 1.

Table 1. The highest accuracy of two models on the CIFAR-10

| Noise Ratio | 10% | 20% | 30% | 40% | 50% |
|---|---|---|---|---|---|
| GBCNN-ResNet | **0.921** | **0.921** | **0.894** | **0.880** | **0.853** |
| CNN-ResNet | 0.892 | 0.878 | 0.860 | 0.840 | 0.810 |

Then, we select two superclasses (vehicles1,vehicles2) with similar attributes in the CIFAR-100 dataset, each with five subcategories, and combine these two superclasses to form a new decile dataset. Furthermore, to prove that the efficient robustness of the granule ball neural network model is not only reflected in the decile class dataset, we selected a superclass (five classifications) on the CIFAR100 dataset for training using the LeNet in order to verify that the model is applicable to any classified dataset and any neural network, as shown in Table 2.

Table 2. The highest accuracy of two kind of models on the subsets of the CIFAR-100

| | CIFAR-100 (18, 19) | | CIFAR-100 (18) | |
|---|---|---|---|---|
| Noise Ratio | GBCNN-ResNet | CNN-ResNet | GBCNN-LeNet | CNN-LeNet |
| 0 | 0.856 | 0.859 | 0.726 | 0.726 |
| 10% | 0.815 | 0.803 | 0.700 | 0.650 |
| 20% | 0.783 | 0.748 | 0.684 | 0.591 |
| 30% | 0.778 | 0.757 | 0.644 | 0.562 |
| 40% | 0.880 | 0.840 | 0.628 | 0.526 |
| 50% | 0.853 | 0.810 | 0.540 | 0.508 |